\documentclass[letterpaper, 10 pt, conference]{ieeeconf}  % Comment this line out if you need a4paper

\IEEEoverridecommandlockouts                              % This command is only needed if 
\usepackage{graphicx} % for pdf, bitmapped graphics files
\usepackage{amsmath} % assumes amsmath package installed
\usepackage{amssymb}  % assumes amsmath package installed
\usepackage{subfig}
\usepackage{siunitx}
\usepackage{xcolor}
\usepackage[ruled,vlined]{algorithm2e}
\usepackage{hyperref}
\newcommand{\Lagr}{\mathcal{L}}

\title{\LARGE \bf
UAV Target Tracking in Urban Environments Using Deep Reinforcement Learning}

\author{Sarthak Bhagat and P.B. Sujit% <-this % stops a space
\thanks{ Sarthak Bhagat is with the Department of Electronics and Communications
Engineering, IIIT-Delhi, India.
{\tt\small sarthak16189@iiitd.ac.in}}%
\thanks{P.B. Sujit is with the Department of Electrical Engineering and Computer Science, Indian Institute of Science Education and Research, Bhopal, India.
{\tt\small sujit@iiserb.ac.in}}%
\thanks{This work is partially supported by EPSRC GCRF Grant EP/P02839X/1}%
}

\begin{document}

\maketitle
\thispagestyle{empty}
\pagestyle{empty}

%%%%%%%%%%%%%%%%%%%%%%%%%%%%%%%%%%%%%%%%%%%%%%%%%%%%%%%%%%%%%%%%%%%%%%%%%%%%%%%%
\begin{abstract}
Persistent target tracking in urban environments using UAV is a difficult task due to the limited field of view, visibility obstruction from obstacles and uncertain target motion. The vehicle needs to plan intelligently in 3D such that the target visibility is maximized. In this paper, we introduce Target Following DQN (TF-DQN), a deep reinforcement learning technique based on Deep Q-Networks with a curriculum training framework for the UAV to persistently track the target in the presence of obstacles and target motion uncertainty. The algorithm is evaluated through several simulation experiments qualitatively as well as quantitatively. The results show that the UAV tracks the target persistently in diverse environments while avoiding obstacles on the trained environments as well as on unseen environments. %Also, our model has the capability to track target in  environments on which it has not trained. % we evaluated our model on dynamic environments also test the ability of our model to cater to dynamic environments and the quantitative results depict that our approach is able to remodel itself effectively when subjected to novel scenarios.
\end{abstract}

%%%%%%%%%%%%%%%%%%%%%%%%%%%%%%%%%%%%%%%%%%%%%%%%%%%%%%%%%%%%%%%%%%%%%%%%%%%%%%%%
\section{Introduction}

% Motivation about target tracking and why it is a challenging problem
Unmanned aerial vehicles (UAVs) have been used for several applications, of which, persistent target tracking is one important application. In a target tracking application, the UAV having a limited  field-of-view (\textit{FOV}) camera needs to continuously track the target. This task becomes a challenge especially in urban environments due to the presence of visibility obstruction from buildings and target motion uncertainty. Therefore, there is a need to develop persistent target tracking algorithms for UAVs in urban environments. 

The target tracking problem has been of interest for over two decades in the robotics literature. One technique to track targets is to develop guidance laws \cite{wise2006uav,choi2014uav,oh2013rendezvous,regina2011uav,chen2009tracking,theodorakopoulos2008strategy,pothen2017curvature}. These guidance laws assume that the target model is known while designing them, satisfying the \textit{FOV} constraints. Moreover, these articles do not consider the presence of obstacles in the environment. Due to the presence of visibility obstruction, one can use multiple cooperative UAVs to meet persistent target tracking \cite{shaferman2008cooperative,cook2013intelligent,yu2014cooperative} requirements. However, we are interested in a single UAV tracking a target in an urban environment with \textit{FOV} constraints, target motion uncertainty and obstacle avoidance.

There have been several efforts in developing vehicle controllers for a single UAV tracking targets in urban environments \cite{zhao2019detection,watanabe2010optimal, semsch2009autonomous,  ramirez2015urban,kim2010uav, wu2018path,theodorakopoulos2009uav}. Zhao {et al.} \cite{zhao2019detection} developed a vision algorithm based on YOLO to detect a target in an urban environment. In this work, a simple proportional controller is used to track the target. However, there are no obstacles in the operational arena. Wantabi and Fabiani \cite{watanabe2010optimal} developed an optimal guidance framework for tracking a target in an urban environment. Although, optimal guidance requires a target model and is also computationally intensive. Semsch et al. \cite{semsch2009autonomous} converted the surveillance  problem to obtain information at certain locations in an urban environment into an art gallery problem taking visibility constraints into account and then use a TSP-based approach to find the path for the UAV. But, in this case, the target tracking aspects were not considered. Ramirez et al. \cite{ramirez2015urban} developed an information-theoretic planner than has an estimate of the target and this estimate is updated based on ground sensors and UAV camera sensor. This work failed to consider obstacles and visibility constraints. Kim and Crassidis \cite{kim2010uav} assigned circular paths to maximize the visibility of the targets and decision to change these circular paths was carried out online using an approximate dynamic programming approach. Wu \textit{et al.} \cite{wu2018path} developed an improved whale optimization framework to determine paths for the UAV to maximize the energy obtained by the solar panels, while considering the presence of obstacles and their shadow into account. They formulated the problem without considering the target tracking aspect. Theodorakopoulos and Lacroix \cite{theodorakopoulos2009uav} developed an iterative optimizing method to track the target in the presence of obstacles. A set of trajectories were predicted and evaluated based on the cost of these trajectories taking the visibility constraint and the obstacle avoidance into account. A path that minimises the cost was determined and given to the UAV for tracking. In our approach, we learn the environment by using reinforcement learning with an unmodeled target motion. Therefore, our approach is robust to different target motion behaviours. 

Learning to track the target is an alternative approach that enables the vehicle to learn the target motion and track it accurately. Zhang et al. \cite{zhang2018coarse} developed a deep reinforcement technique for a camera to track the target under different aspect ratios. However, this approach does not consider urban environments. Mueller et al. \cite{mueller2016benchmark} developed a UAV target tracking simulator where one can evaluate different computer vision algorithms for target detection and tracking offline and also developed UAV tracking controllers for different target trajectories. The simulator utilized in this work had a limitation on the urban environment setting. Learning-based controllers for UAV target tracking in urban environments have been inadequately studied. We present a deep reinforcement learning technique called Target Following Deep Q-Network (TF-DQN) for persistent target tracking in urban environments. We use a DQN because of its simplicity of implementation as the UAV control is discretized. Moreover, DQN requires less training data, do not exhibit the overestimation property \cite{Fujimoto2018AddressingFA} and the convergence in the training of DQN and its derivatives like DDQN \cite{Hasselt2015DeepRL} is stabler than policy gradient methods. Additionally, one key drawback of policy gradient methods is that they suffer from high variance in estimating the gradient of $\mathbb{E}[R_{t}]$, which is not the case in a DQN.

Our main contributions can be summarised as follows:
\begin{itemize}
    \item We propose a Deep Q-Network based approach to persistently track a dynamic target.
    \item We develop a simulator that mimics an urban environment with plausibly dynamic obstacles and a target vehicle that can move with varying velocity  in the environment.
    \item We test our proposed approach using qualitative as well as quantitative evaluation that includes a set of three diverse metrics to quantify its efficacy.
    \item Lastly, we test our model's ability to adapt to changes in the environment's configuration via curriculum training.
\end{itemize}

The rest of the paper is organized as follows. The problem formulation is described in  Section \ref{sec:pf} and the entire framework of TF-DQN is described in Section \ref{sec:deep}. Evaluation of the proposed approach is carried out in Section \ref{sec:simulations} and we conclude in Section \ref{sec:conclusions}. 

\section{Problem Formulation}\label{sec:pf}
\begin{figure}
	\centering
	\includegraphics[scale=0.4]{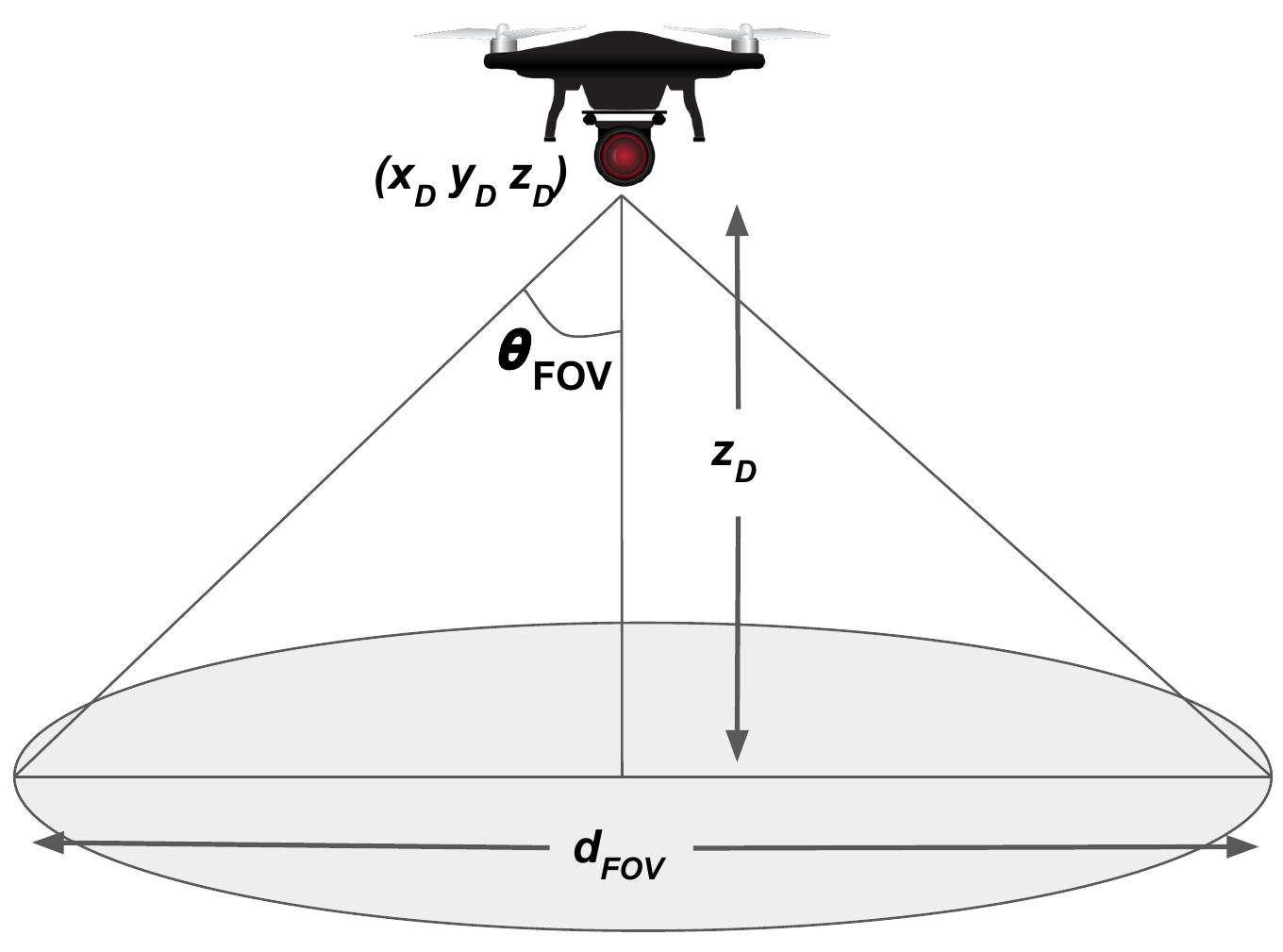}
	\caption{An illustration of the field of view (FOV) of a UAV placed at $(x_{D}, y_{D}, z_{D})$ and having a maximum viewing angle of $\theta_{FOV}$.}
	\label{fig:fov}
\end{figure}

We consider a scenario where a target is moving in an urban environment consisting of obstacles and a road network. The obstacles in the environment are represented as cylindrical objects of varying sizes (radius and height). We consider cylindrical obstacles for simplicity. While maneuvering, when a target reaches a junction (meeting of two or more roads), it can randomly select any road segment. The UAV has a fixed-sized field of view (represented as \textit{FOV}, see Figure \ref{fig:fov}) which is a function of its altitude. The objective of the UAV is to persistently track the target by ensuring the target is within its \textit{FOV}. 

We define the position of UAV as $p_{D} = (x_{D}, y_{D}, z_{D})$ and position of target as $p_{T} = (x_{T}, y_{T})$. Due to the camera resolution, we assume that the vehicle can changes its altitude between  $z_{D} \in [h_{D}^{min}, h_{D}^{max}]$.
We also define $n \in N$ cylindrical obstacles with the $i^{th}$ obstacle having radius $r_{i}$, height $h_{i}$ and center located at $(x_{O_{i}}, y_{O_{i}})$. We represent the entire state space and the action space of the UAV as $\mathbb{S}$ and $\mathbb{A}$ respectively.

We define a binary variable $C^{i}$ that is ${\tt true}$ (or 1) if the UAV collides with the $i^{th}$ obstacle and ${\tt false}$ (or 0) otherwise.
\begin{equation}
    C^{i}(p_{D}, p_{i}) =    \left\{
                    \begin{array}{ll}
                      1 & \sqrt{(x_{D} - x_{O_{i}})^{2} + (y_{D} - y_{O_{i}})^{2}} \leq r_{i}, \\ 
                      & z_{D} \leq h_{i} \\
                      0 & otherwise. \\
                \end{array} 
                \right. 
\end{equation}

Equation of the line joining the UAV and the target at a given time step is given by Equation \ref{equation_of_line}.
\begin{equation}
\label{equation_of_line}
    \frac{x - x_{D}}{x_{T} - x_{D}} = \frac{y - y_{D}}{y_{T} - y_{D}} = \frac{z - z_{D}}{- z_{D}}.
\end{equation}

The binary variable $I^{i}$ is ${\tt true}$ (or 1) if the line joining the UAV and the target intersects with the $i^{th}$ obstacle and ${\tt false}$ (or 0) otherwise. On solving Equation \eqref{equation_of_line}, we get the following condition for $I^{i}$.

\begin{equation}
    I^{i}(p_{D}, p_{T}, p_{i}) =    \left\{
                    \begin{array}{ll}
                      1 & \frac{z_{D}(-x_{O_{i}} + x_{D})}{x_{T} - x_{D}} + z_{D} \leq h_{i}, \\
                      & \frac{(x_{T} - x_{D})y_{O_{i}} + (y_{T} - y_{D})x_{O_{i}}}{\sqrt{(x_{T} - x_{D})^{2} + (y_{T} - y_{D})^{2}}} \leq r_{i} \\
                      0 & otherwise \\
                \end{array} 
                \right. 
\end{equation}

We also define another binary variable $V(p_{D}, p_{T})$ that represents the visibility state of the target. It is ${\tt true}$ (or 1) if the target lies in the FOV of the UAV and ${\tt false}$ (or 0) otherwise.
\begin{equation}
    V(p_{D}, p_{T}) = \left\{
                    \begin{array}{ll}
                      1 & \Big(x_{D} + \frac{d_{FOV}}{2}\Big) \leq  x_{T} \leq \Big(x_{D} + \frac{d_{FOV}}{2}\Big), \\
                      & \Big(y_{D} + \frac{d_{FOV}}{2}\Big) \leq  y_{T} \leq \Big(y_{D} + \frac{d_{FOV}}{2}\Big) \\
                      0 & otherwise \\
                \end{array} 
                \right.
\end{equation}
where, the diameter of the field of view of the UAV $d_{FOV}$ is a function of the altitude $z_{D} \in [h_{D}^{min}, h_{D}^{max}]$ of the UAV and is given by
\begin{equation}
    d_{FOV} = 2 z_{D} \tan(\theta_{FOV}) ,
\end{equation}
where, $\theta_{FOV}$ is the maximum possible viewing angle of the UAV calculated from the perpendicular drawn from its position to its projection on the x-y plane.

\section{Deep Reinforcement Learning Method}\label{sec:deep}

We introduce a target tracking technique that sequentially builds upon its prior knowledge of the environment as well as the current location of the target. This novel approach called as Target Following Deep Q-Network (TF-DQN) uses deep reinforcement learning techniques to address different constraints and objectives of the problem.

\subsection{Q-Learning}

At each time step $t$, the state of the agent is represented as $s_{t} = p_{D} = (x_{D}, y_{D}, z_{D}) \in \mathbb{S}$. The agent chooses an action $a_{t} \in\mathbb{A} = \{1, 2, .. |A|\}$ receiving a reward $r_{t}$. 

The aim of the agent is to maximise the expected discounted return which is given by $R_{t} = \sum_{\tau=1}^{\infty} \gamma^{\tau - t}r_{\tau}$ where $\gamma \in [0, 1] $ is the discount factor. The value of each state action pair $(s_{t}, a_{t})$ also referred to as the Q-function is represented in the form of an expectation function as
\begin{equation}
    Q^{\pi}(s_{t}, a_{t}) = \mathbb{E}\Big[R_{t} | s = s_{t}, a = a_{t}, \pi\Big],
\end{equation}
where $\pi$ represents the policy followed by the agent. 

Using a dynamic programming based approach and Bellman's optimality equation, we obtain the optimal Q-function as
\begin{equation}
    Q^{*}(s_{t}, a_{t}) = \mathbb{E}_{s'}\Big[{r + \gamma \max_{a'}Q^{*}(s', a') | s, a}\Big].
\end{equation}
Hence, this model-free reinforcement learning algorithm generates an optimal policy for the agent by maximizing the expected value of the total return beginning from a given state $s$ and determining a specific action $a$.

\subsection{Deep Q-Networks}
The Q-functions elaborated in the previous section represent the values of each state-action pair for a particular configuration of the environment. For depicting a function with a large number of such state-action pairs possible, Deep Q-Networks (DQNs) utilize the function approximation property of neural networks to represent Q-functions as a function of the weight parameter $\theta$ as $Q(s, a; \theta)$. Here, $s$ and $a$ represent the state and action respectively while the next state (obtained on performing $a$ at $s$) is $s'$.

At each iteration $i$, the agent's state and action is fed to the DQN and the estimated Q-value is generated. We attempt to make our prediction as close as possible to the target value which is given by
\begin{equation}
    y_{i}^{DQN} = r + \gamma \max_{a'}Q(s', a'; \theta^{-}),
\end{equation}
by optimising the loss function represented in the form of an expectation over $s, a, r, s'$ given by
\begin{equation}
    \label{dqn_loss}
    \Lagr_{i}(\theta_{i}) = \mathbb{E}_{s, a, r, s'}\bigg[\frac{1}{2}\bigg(y_{i}^{DQN} - Q(s, a; \theta_{i})\bigg)^{2}\bigg],
\end{equation}
where $\theta^{-}$ denotes  the target network parameters. The two key findings of \cite{Mnih2013PlayingAW} include

\begin{itemize}
    \item \textit{Target Network:} For improving the stability of training by ensuring the predictions do not follow a moving target, \cite{Mnih2013PlayingAW} proposed freezing of weights of a target network with parameter $\theta^{-}$ for a fixed number of $\tau$ iteration while updating the weights of the online network. After every $\tau$ iterations, the weights of the target network are equated with that of the online network.
    \item \textit{Experience Replays:} During training, experiences $e_{t} = (s_{t}, a_{t}, r_{t}, s_{t+1})$ are collected over many episodes to form a dataset of experiences $\mathcal{D} = \{e_{1}, e_{2}, e_{3} ... |\mathcal{D}| \}$. Instead of using samples in the standard sequence, mini-batches are randomly sampled from this dataset for training in order to reduce correlation between samples. 
\end{itemize}

Including these in Equation \eqref{dqn_loss} gives 
\begin{equation}
    \Lagr_{i}(\theta_{i}) = \mathbb{E}_{(s, a, r, s')\sim \mathcal{D}}\bigg[\frac{1}{2}\bigg(y_{i}^{DQN} - Q(s, a; \theta_{i})\bigg)^{2}\bigg],
\end{equation}
whose gradient is given by
\begin{align}
    \nabla_{\theta_{i}} \Lagr_{i}(\theta_{i}) = \mathbb{E}_{(s, a, r, s')\sim \mathcal{D}}\bigg[\bigg(y_{i}^{DQN} - Q(s, a; \theta_{i})\bigg) 
    \notag \\ 
    \nabla_{\theta_{i}}Q(s, a; \theta_{i}\bigg].
\end{align}

This gradient is used to update $\theta$ employing a gradient descent algorithm which in turn is required to obtain the optimal set of parameters for the model. The learning rate for the weight update is gradually decreased over epochs to ensure the parameters converge to the optimal set of parameters.

Various extensions of DQN like Double Q-Learning \cite{Hasselt2015DeepRL}, Dueling Network \cite{Wang2015DuelingNA}, Bootstrapped DQN \cite{Osband2016DeepEV}, Deep Recurrent Q-Network (DRQN) \cite{Hausknecht2015DeepRQ}, etc. have been instrumental in the development of perception and control using RL. 

\subsection{Target Following DQN}
\begin{figure*}
      \centering
      \includegraphics[scale=0.7]{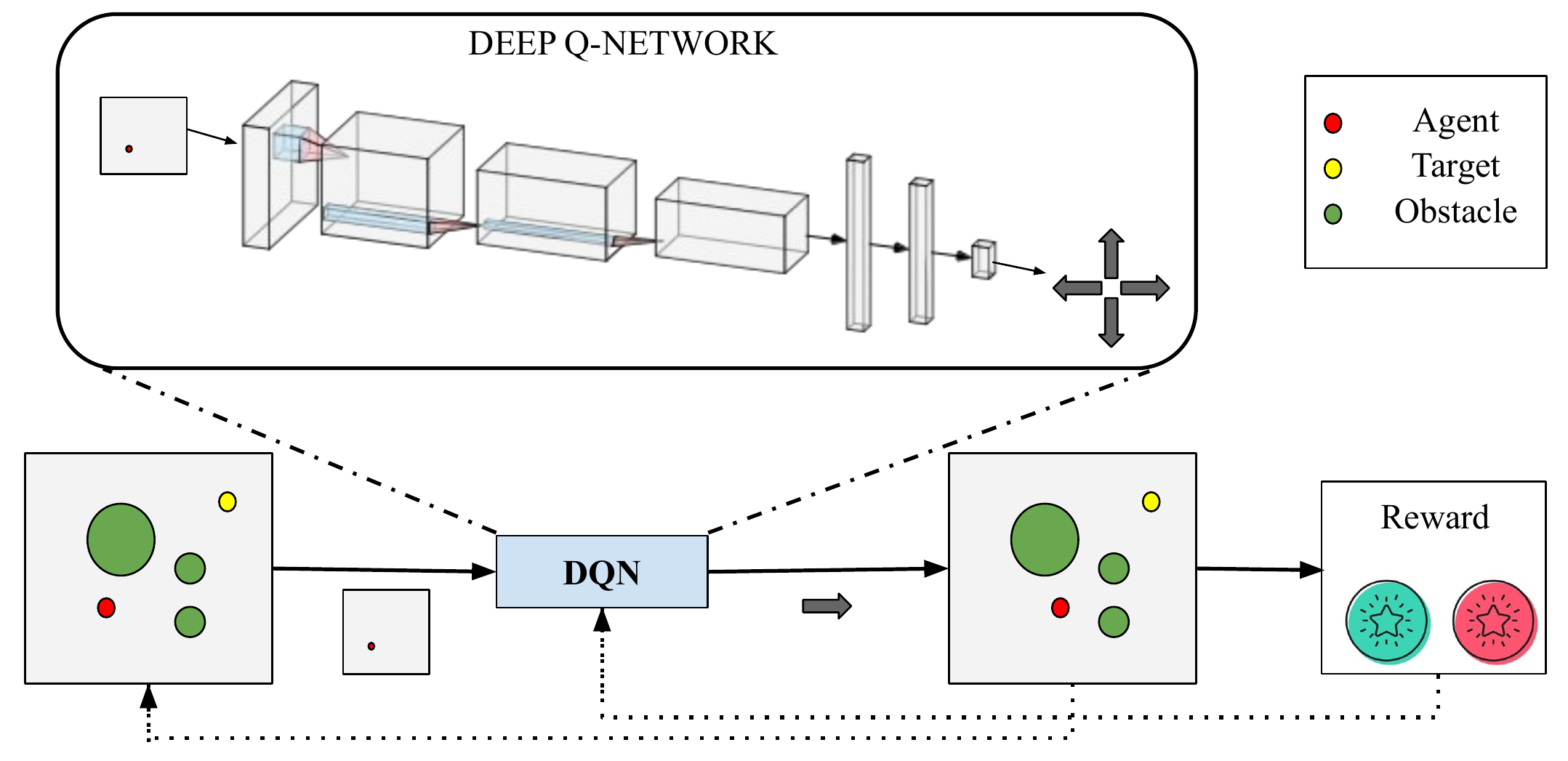}
      \caption{Training procedure for Target Following DQN (TF-DQN). The state of the agent is passed on to the DQN which in turn returns the action to be performed. The reward received by the agent on performing this particular action is utilized to update the Q-function of this state-action pair.
      The dotted lines in the figure depict the update steps in the implementation.}
      \label{fig:approach}
   \end{figure*}
Target Following DQN (TF-DQN) is a novel target tracking technique that is specialized for tracking a mobile ground vehicle using a UAV in a dynamic environment that consists of obstacles, target motion uncertainty and occlusions.

\subsubsection{Exploration vs Exploitation}
For designing this algorithm, at each state $s_{t} \in \mathbb{S}$, the agent picks an action $a_{t} \in \mathbb{A}$ using a probabilistic approach in which the agent tends to explore more in the beginning and gradually moves to a policy that predominantly exploits the learned Q-function to pick the most rewarding action.

The approach utilized to pick the action at each step of the $k^{th}$ episode is as follows
\begin{equation}
    a_{t} = \left\{
                    \begin{array}{ll}
                      rand(A) & (1 - p_{sat})e^{-\alpha k} + p_{sat} \\ 
                      \underset{a} {\mathrm{argmax}} ~Q(s, a) & otherwise \\
                \end{array} 
                \right. 
\end{equation}
where $p_{sat}$ is the saturation probability and $\alpha$ is a positive constant. So, using the given approach our model picks only random actions in the beginning and later as episodes progresses, stabilizes to a constant exploration probability $p_{sat}$.

\subsubsection{Piecewise Reward Function}

We obtain the reward received at each time step for a given state-action pair using the Algorithm \ref{algo:reward}. At each time step, this algorithm takes in as input the state of the individual elements in the environment and outputs the instantaneous reward by investigating the possible collision with the obstacle, obstruction of the target due to one (or more) of the obstacles, and the visibility of the obstacle within the \textit{FOV} of the agent. 

 The reward $r_{t}$ at any time step $t$ is given by
 \begin{equation}
     r_{t} = \left\{
                     \begin{array}{ll}
                       R_{c} & C^{i}(p_{D}, p_{i}) = 1; i \in [1, n]  \\ 
                       R_{i} & I^{i}(p_{D}, p_{T}, p_{i}) = 1; i \in [1, n] \\
 R_{v}(p_{D}, p_{T}) & V(p_{D}, p_{T}) = 1 \\
                       R_{nv}e^{-\beta t_{nv}} & V(p_{D}, p_{T}) = 0. \\
                 \end{array} 
                 \right.
                 \label{eq:reward_eq}
 \end{equation}
where, $R_{c}$ is the collision reward constant, $R_{i}$ is the intersection reward constant, $R_{v}(p_{D}, p_{T})$ is the positive reward function and $R_{nv}$ is the negative reward constant. Here, $t_{nv}$ is the time corresponding to target invisibility and is reset each time the agent views the target. This variable is incremented at each time step the agent is unable to view the target. 
The collision reward constant is kept as a large negative value that ensures the agents stay away from the obstacles while maneuvering.

We also observe that there are situations when the agent gets stuck around an obstacle while the target is on the other side of it adhering to the fact that it cannot cross the obstacle due to the hugely negative collision reward. To avoid such a situation, we ensure that the agent receives a negative intersection reward, $R_{i}$ when the line joining it to the target intersects with any of the obstacles. This prevents the obstructions of the view of the agent due to the obstacles present in the environment.

When $V(p_{D}, p_{T}) = 0$, the negative loss which is an exponential function of $n_{nv}$ enforces the agent to keep the target in its \textit{FOV} by penalising not tracking it for a long duration of time.

The positive reward function $R_{v}(p_{D}, p_{T})$ is given by:
\begin{equation}
    R_{v}(p_{D}, p_{T}) = \frac{R_{v}^{c}}{\sqrt{(x_{D} - x_{T})^{2} + (y_{D} - y_{T})^{2}}} + \frac{h_{v}^{c}}{z_{D}},
\end{equation}
where $R_{v}^{c}$ and $h_{v}^{c}$ are positive constants.

The first term of the positive reward function $R_{v}(p_{D}, p_{T})$ encourages the agent to align itself in such a way that the target is as close to the center of the \textit{FOV} of the agent as possible. While the second term encourages it to keep its altitude as low as possible while trying to capture the view of the target precisely.

\begin{algorithm}
\SetAlgoLined
\KwResult{Reward $r_{t}$ received by the agent at time step $t$ for picking an action $a_{t}$ at state $s_{t}$.}
\textbf{Input:} Current state of agent $p_{D}$, current state of target $p_{T}$, state of each obstacle $p_{i} \forall i \in [1, n]$, maximum allowed length of each episode $t_{max}$. Reward constants: $R_{c}$, $R_{i}$, $R_{v}^{c}$, $h_{v}^{c}$, $R_{nv}$ and $\beta$.

Initialize $t$ = 0\;

 \While{$t \leq t_{max}$}{
    
    $collision$ = {\tt false}\;
    $intersection$ = {\tt false}\;
    
    \For{$i\gets1$ \KwTo $n$}{
        \eIf{$C^{i}(p_{D}, p_{i}) = 1$}{
           $collision$ = {\tt true}\;
           $break$\;
           }{}
        \eIf{$I^{i}(p_{D}, p_{T}, p_{i}) = 1$}{
           $intersection$ = {\tt true}\;
           }{}
    }
    
    \eIf{$collision$ = {\tt true}}{
       $t_{nv} \gets t_{nv} + 1$\;
       return $R_{c}$;
       }
       {
       \eIf{$intersection$ = {\tt true}}{
       $t_{nv} \gets t_{nv} + 1$\;
       return $R_{i}$\;
       }
       {
       \eIf{$V(p_{D}, p_{T}) = 1$}{
       $t_{nv} = 0$\;
       return $R_{v}(p_{D}, p_{T})$\;
       }
       {
       $t_{nv} \gets t_{nv} + 1$\;
       return $R_{nv}e^{-\beta t_{nv}}$\;
      }
      }
   }
   $t \gets t + 1$\;
 }
 \caption{Obtaining reward $r_{t}$ at each state-action pair trying to follow a maneuvering target.}
 \label{algo:reward}
\end{algorithm}
\subsubsection{Exploration in the Search-Space}

Another key feature of TF-DQN is the exploration in the uncertain regions that prevents the agent from local minima trap in some regions away from the target due to a sub-optimal learned policy. During training, the agent often reaches regions close to the environment boundaries when it is not able to view the target for a long duration of time and therefore, does not receive a reward. This leads to an almost static Q-function for this duration and the agent fails to move towards an optimal policy. Such a situation also leads to an unbalanced and unexplored environment and therefore, the Q-function corresponding to these regions is unrepresentative of the {\tt true} value of the state-action pair. 

In order to mitigate the mentioned drawbacks, we ensure that the agent sufficiently explores while the target is not visible for a certain duration of time. The agent in such a situation enters into the Search-Space where it predominantly explores by picking a random action and occasionally picks the optimal action according to the Q-function with a much lower probability. The agent enters into this space only when the difference between the current time step and the last time agent was able to capture the target ($t_{nv}$) is above a certain threshold i.e. $t_{nv} \geq t_{nv}^{threshold}$. While the agent is in the Search-Space, the action $a_{t}$ picked by the agent at state $s_{t}$ is given by

\begin{equation}
    a_{t} = \left\{
                    \begin{array}{ll}
                      rand(a) & p_{SS} \\
                      \underset{a} {\mathrm{argmax}} ~Q(s, a) & 1 - p_{SS} \\
                \end{array} 
                \right.
\end{equation}
where, $p_{SS} \approx 1 $. 

\subsubsection{Lifelong and Curriculum Learning}
Due to the dynamic nature of this task of following a given target and avoiding obstacles in the path, we ensure that our models never stops learning and rather utilizes the understanding of the environment based on the data already seen to predict the optimal actions in the current time step. Our model progressively builds on the understanding of the environment and the current location of the target and therefore, is generalizable to newer scenarios and variations in the setup. The learning rate gradually reduces to a stable final value to avoid any sudden fluctuations in the weight parameters.

Our model is capable of handling multiplex environments with varying obstacle configurations by utilizing the already acquired information about the surroundings. Fine-tuning the Q-function for the newer set of obstacles enables us to learn the optimal set of parameters in a much smaller amount of time to perform equivalently well even if it had been trained on this environment itself. 

\section{Simulation Results}\label{sec:simulations}

\subsection{Urban Environment Simulator}
\begin{figure}
      \centering
      \includegraphics[scale=0.25]{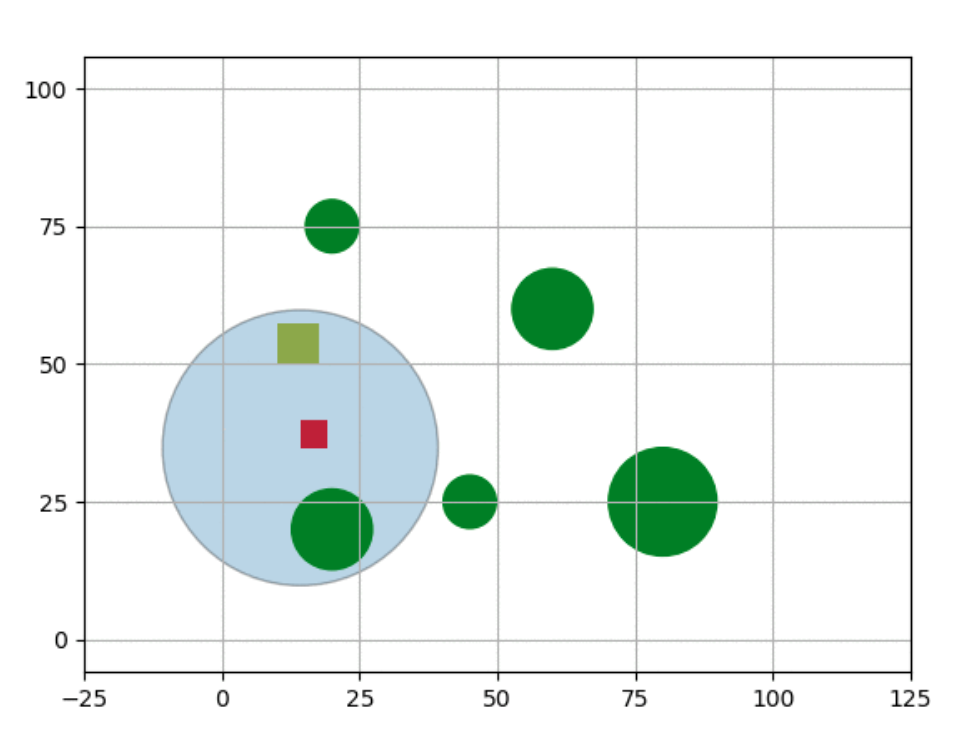}
      \caption{A screenshot of the simulator environment with red square depicting the UAV, yellow square depicting the target vehicle, green circles depicting obstacles and the large blue circle depicting the FOV of the UAV. The road network for the target vehicle to maneuver is depicted by the black lines in the figure.
      The environment available to the target is in the shape of a square with side $s = 100$.}
      \label{fig:simulation_env}
   \end{figure}

We design a simulator \footnote{Code for this simulator is available at \href{https://github.com/sarthak268/Target-Tracking-Simulator}{https://github.com/sarthak268/Target-Tracking-Simulator}} to replicate the real-world urban environment with a UAV, a mobile ground vehicle that acts as the UAV's target and some cylindrical obstacles. A screenshot of the simulator is shown in Figure \ref{fig:simulation_env}.

The UAV can pick an action from a domain of 6 possible actions which gives us $A = \{{\tt north}, {\tt south}, {\tt west}, {\tt east}, {\tt up}, {\tt down}\}$. The environment is of the form of a square with each side as $s$, beyond which the target vehicle does not move, while the UAV is free to move anywhere until its altitude lies within the range $[h_{D}^{min}, h_{D}^{max}]$. Additionally, we assume that there exists only a certain discrete number of height levels ($n_{h}$) that the UAV can take rather than a continuous spectrum of altitude values. These height levels are equally spaces between $h_{D}^{min}$ and $h_{D}^{max}$, therefore, an increment in each height level results in an increase in height by $h_c$ i.e. $(h_{D}^{max} - h_{D}^{min}) = n_{h} h_{c}$.

The target vehicle randomly chooses a direction at each junction of the road and therefore, its movement is probabilistic. 
We also keep the speed of the drone constant and greater than or equal to that of the target vehicle in simulation experiments.

\subsection{Results}

\subsubsection{Implementation Details}

For the construction of the Deep Q-Network, we use 3 convolutional layers followed by a couple of linear layers. Between each layer, we introduce non-linearity using the ReLU activation function and Instance Normalisation between each convolutional layer. This network takes the state of the agent as input and returns the action for the agent to pick. We also list the range of all the hyperparameters used for conducting the simulation experiments in Table \ref{table:hyperparameters}.

\begin{table}
    \caption{Hyperparameters used in the simulation experiments}
    \label{table:hyperparameters}
    \begin{center}
        \begin{tabular}{|p{5cm}|c|c|}
        \hline
        \textbf{Hyperparameter Name} & \textbf{Symbol} & \textbf{Range} \\
        \hline
        Collision Reward Constant & $\|R_{c}\|$ & 1000-2000 \\
        \hline
        Intersection Reward Constant & $\|R_{i}\|$ & 30-100 \\
        \hline
        Positive Reward Distance Constant & $R_{v}^{c}$ & 3000-4500 \\
        \hline 
        Positive Reward Height Constant & $h_{v}^{c}$ & 1500-5000 \\
        \hline
        Negative Reward Constant & $\|R_{nv}\|$ & 1-50 \\
        \hline
        Time Constant in Negative Reward & $\beta$ & 1-10 \\
        \hline 
        Episode Constant in Action Selection & $\alpha$ & 0.1-5 \\
        \hline
        Saturation Probability & $p_{sat}$ & 0.1-0.4 \\
        \hline
        Search-Space Probability & $p_{SS}$ & 0.9-0.95 \\
        \hline 
        Min. Attainable Height for UAV & $h_{min}$ & 1-10 \\
        \hline
        Max. Attainable Height for UAV & $h_{max}$ & 10-60 \\
        \hline
        Threshold Steps for entering Search-Space & $t_{nv}^{threshold}$ & 3-10 \\
        \hline
        Num. of Obstacles in the Environment & $n$ & 2-7 \\
        \hline
        Side of the Square of Environment & $s$ & 100-200 \\
        \hline
        Num. of Possible Height Levels for the UAV & $n_{h}$ & 5-20 \\
        \hline
        Height of obstacles & $h_{i} \forall i \in n$ & 1-50 \\
        \hline
        Height Constant & $h_{c}$ & 1-10 \\
        \hline
        Radius of obstacles & $r_{i} \forall i \in n$ & 2.5-10 \\ \hline
        Maximum Viewing Angle of UAV & $\theta_{FOV}$ & $\ang{30}$-$\ang{45}$ \\
        \hline
        Maximum length of an episode & $t_{max}$ & 500 \\
        \hline
        Discount Factor in Return Calculation & $\gamma$ & 0.1 \\
        \hline
        Learning Rate in Gradient Descent & $\alpha_{LR}$ & 0.01 \\
        \hline
        \end{tabular}
    \end{center}
\end{table}

\subsubsection{Quantitative Results}
We evaluate the efficacy of our approach using three diverse metrics.
\begin{itemize}
    \item \textit{Avg. Distance.} The average distance between the UAV and the target should decrease over time as the UAV should be able to follow the target more closely.
    \item \textit{Avg. Time.} The average time for which the target is visible to the UAV should increase over time as the UAV should be able to view the target for a longer duration of time. This can take a maximum value of 500 as we clamp the maximum length of an episode to this value. 
    \item \textit{Avg. Reward.} The average reward should increase over time as the UAV would ensure the target remain as close to it as possible and therefore, receiving a higher reward.
\end{itemize}

\begin{figure}
      \centering
      \includegraphics[scale=0.25]{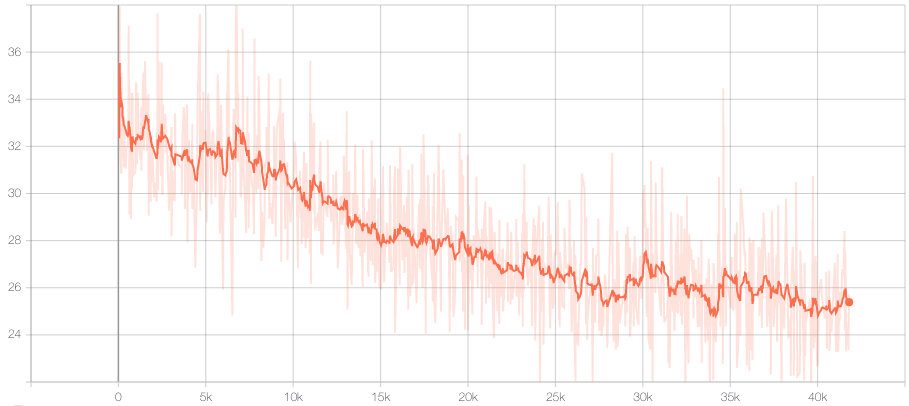}
      \caption{Training Plot for \textit{Avg. Distance}.}
      \label{fig:avg_distance}
   \end{figure}
   
\begin{figure}
      \centering
      \includegraphics[scale=0.25]{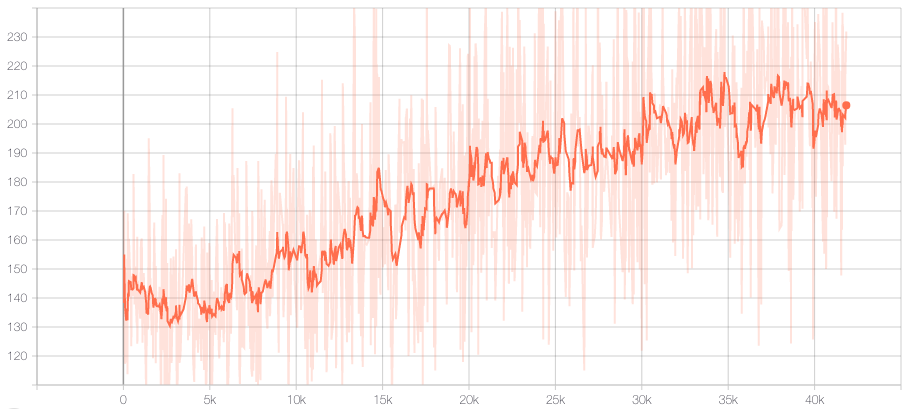}
      \caption{Training Plot for \textit{Avg. Time}.}
      \label{fig:avg_visible_time}
   \end{figure}

\begin{figure}
      \centering
      \includegraphics[scale=0.25]{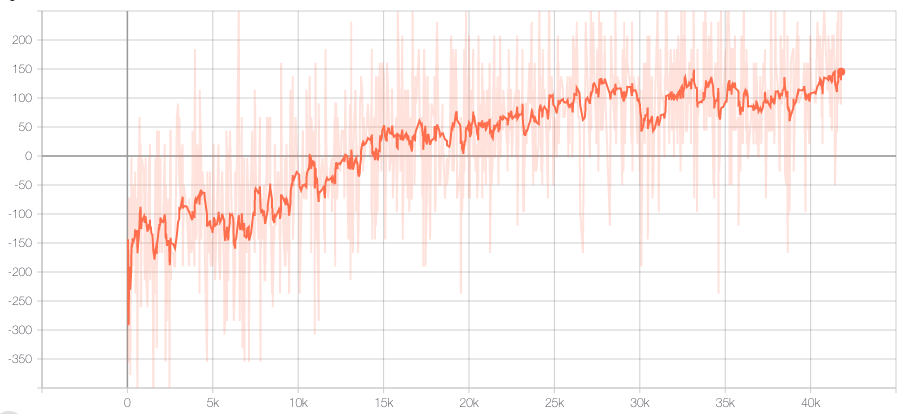}
      \caption{Training Plot for \textit{Avg. Reward}.}
      \label{fig:avg_reward}
   \end{figure}

On plotting these metrics while training in plots \ref{fig:avg_distance}, \ref{fig:avg_visible_time} and \ref{fig:avg_reward}, we observe a significant improvement in performance that justifies our approach of gradually learning the expected value of the state-action pairs in order to follow a mobile target. The plots clearly depict the progressive elevation of the average reward received by the agent and the average duration for which the target lies in the \textit{FOV} of the agent alongside the demotion of the average distance between the target and the agent. Our approach of utilizing a DQN for learning the average return for state-action pairs by penalising not being able to persistently track the target and shifting to a primarily exploratory policy ensures that our agent always continues learning the optimal policy. 

\textit{Curriculum Training Inference.} We also evaluate the performance of our approach to dynamic environments. After sufficient training, we vary the number, position, and size of obstacles gradually in the environment and compute these three metrics after minor fine-tuning on the newly obtained surroundings. This tests the ability of our approach to cater to real-world scenarios where the environment is constantly evolving, and the agent should continue learning by utilizing the knowledge of the past to maximize its reward in the future. In this experiment, the position and size of obstacles in the environment for a particular $n$ are kept fixed for a fair comparison.

\begin{table}
    %\label{table:dynamic_env}
    \begin{center}
        \begin{tabular}{|c|c|c|c|}
        \hline
        \textbf{Num. of Obstacles, $n$} & \textbf{\textit{Avg. Distance}} & \textbf{\textit{Avg. Time}} & \textbf{\textit{Avg. Reward}} \\
        \hline
        3 & 24.6 & 205.4 & 160.4 \\
        \hline
        5 & 28.7 & 184.0 & 112.2 \\
        \hline
        7 & 34.2 & 143.2 & 72.2 \\
        \hline
        3 $\rightarrow$ 5 & 29.4 & 176.8 & 106.7 \\
        \hline
        5 $\rightarrow$ 7 & 36.5 & 137.7 & 68.4 \\
        \hline
        3 $\rightarrow$ 7 & 43.6 & 128.2 & 62.8 \\
        \hline
        \end{tabular}
    \end{center}
    \caption{Performance of our approach on varied number on obstacles and on the task of Curriculum Training. Here, $a$ $\rightarrow$ $b$ represents a situation wherein a model that is trained on $a$ obstacles is fine-tuned for $b$ obstacles.} \label{table:curriculum}
\end{table}

As shown in Table \ref{table:curriculum}, our model is able to adapt sufficiently well to unseen environments and is able to achieve a performance that is close to that achieved when directly trained in those environments. The ability of our model to adapt to novel scenarios and achieve a satisfactory performance with minor fine-tuning makes it computationally efficient for training target-following agents for a wide range of settings. In the real-world, it is not feasible to train an agent for all possible scenarios (due to the evolving nature of the environment), therefore, models that are able to customize themselves to newer assignments hold utmost importance.

\subsubsection{Qualitative Results}

\begin{figure*}
      \centering
      \subfloat[Example 1]{\includegraphics[scale=0.65]{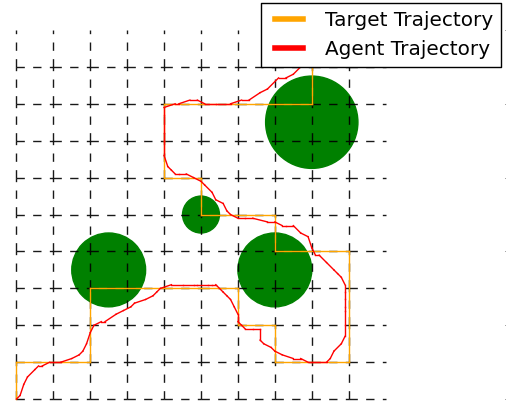}}
      \subfloat[Example 2]{\includegraphics[scale=0.65]{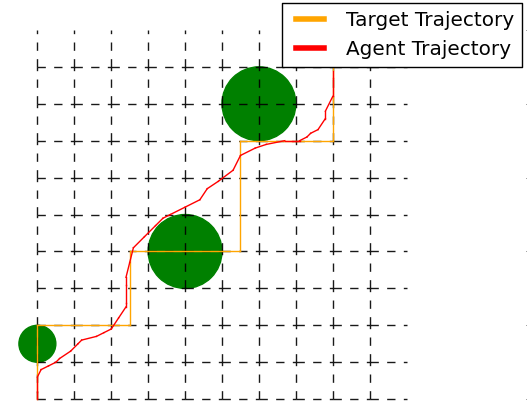}}\\
      \subfloat[Example 3]{\includegraphics[scale=0.65]{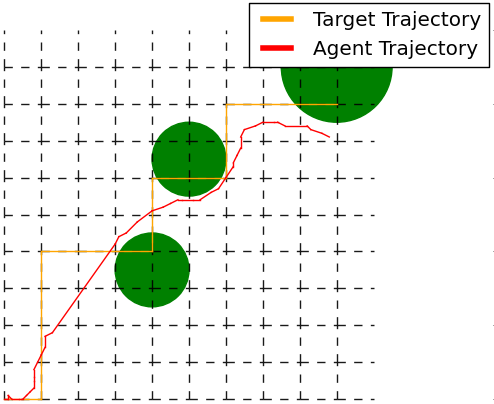}}
      \subfloat[Example 4]{\includegraphics[scale=0.65]{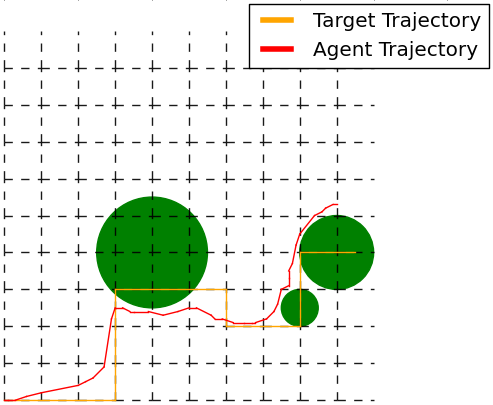}}
      \caption{Trajectory of the UAV and the target vehicle when the maximum height attainable for the drone is lower than the height of any of the obstacles i.e. $h_{D}^{max} \leq h_{i} \forall i \in n$.}
      \label{fig:trajectory_without_height}
   \end{figure*}
   
\begin{figure*}
      \centering
    \subfloat[]{\label{fig:trajectory_with_height_1}  \includegraphics[scale=0.7]{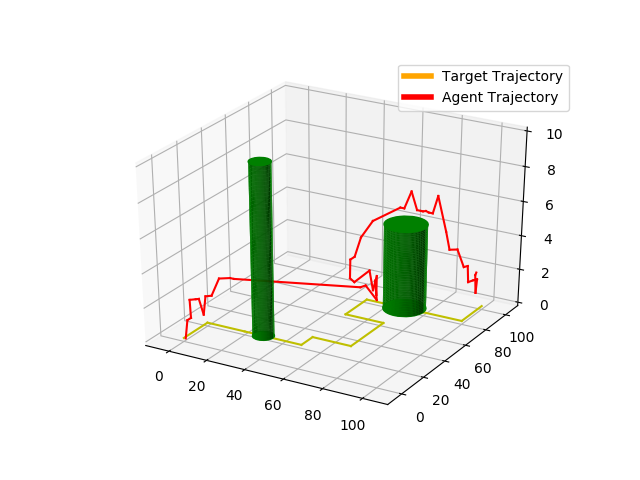}}
  \subfloat[]{ \label{fig:trajectory_with_height_2d_1}     \includegraphics[scale=0.55]{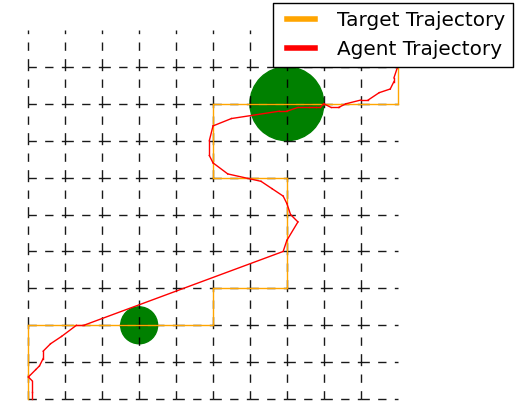}}
            \caption{(a) 3-Dimensional Trajectory of the vehicle (b)  2-dimensional projection on the x-y plane of the UAV and the target vehicle when the maximum height attainable for the drone is higher than height of atleast one of the obstacles i.e. $h_{D}^{max} \geq h_{i}$ for some $i \in n$.}
   \end{figure*}

\begin{figure*}
      \centering
    \subfloat[]{\label{fig:trajectory_with_height_2}  \includegraphics[scale=0.7]{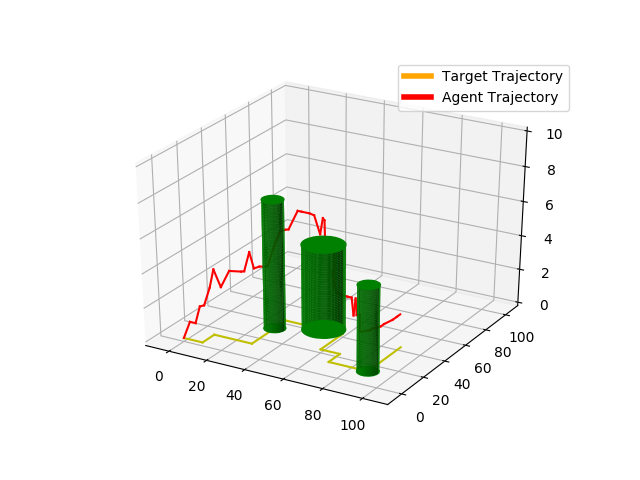}}
  \subfloat[]{ \label{fig:trajectory_with_height_2d_2}     \includegraphics[scale=0.7]{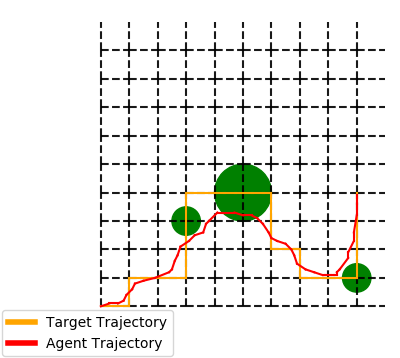}}
            \caption{(a) 3-Dimensional Trajectory of the vehicle (b)  2-dimensional projection on the x-y plane of the UAV and the target vehicle when the maximum height attainable for the drone is higher than height of atleast one of the obstacles i.e. $h_{D}^{max} \geq h_{i}$ for some $i \in n$.}
   \end{figure*}

We also evaluate the performance using a qualitative approach by plotting the trajectory of the UAV as well as the target vehicle on the map of the environment. 
The plots of the trajectory provided represent the path followed by the target vehicle after sufficient training alongside the path that the UAV follows in order to track the target. The red curve represents the UAV trajectory, the orange curve represents the target's trajectory and the green cylinders represent the obstacles in the environment. The dotted lines represent the road network where the target can maneuver.

Figure \ref{fig:trajectory_without_height} depicts situations in which $h_{D}^{max} \leq h_{i} \forall i \in n$, therefore, varying the height of the UAV does not aid it to avoid the obstacles. In these cases, the UAV has to altogether avoid the obstacles in the x-y plane itself. The exceedingly large negative reward received on collision enforces the agent to avoid them, but this does not interfere in the prime task of following the target. 

Figures \ref{fig:trajectory_with_height_1} and \ref{fig:trajectory_with_height_2} depict 3-dimensional plots of the trajectories of the UAV and target where the UAV also varies its altitude to avoid the obstacles by increasing the altitude. So, in order to maximize its reward, it chooses to avoid some obstacles by going around it (see Figure \ref{fig:trajectory_with_height_2d_1} and \ref{fig:trajectory_with_height_2d_2}) while the others by going over it. The ability of our model to not only search the 2D space but also vary its altitude in order to maximise its reward, ensures that it closely replicates the real-world scenario. 

\section{Conclusions}\label{sec:conclusions}

In this paper, we proposed TF-DQN, a target tracking deep reinforcement learning based approach, that suggests actions for the UAV to track a maneuvering ground vehicle persistently in a complex environment. The proposed approach is computationally simple and easy to implement. The simulation results show that the vehicle is able to track the target in 3D persistently besides avoiding obstacles in its path. 

Next, we would like to experimentally validate the approach on a quadrotor with a simple vision-based system to detect the target. This work can be extended for multiple UAV agents to follow multiple targets by collaborating with each other. Due to the presence of multiple agents learning becomes a complex task. Another possible direction is to extend the approach for fixed-wing UAV which poses kinematic constraints.   

%Multi-target tracking \cite{vo} can be another interesting direction of work wherein the aim of a swarm of UAVs is to track down a collection of targets. This is intrinsically an NP-hard problem and therefore, a reinforcement learning based approach could be more effective in computing the optimal action at each state as compared to completely heuristic approach.
%We intend to continue testing our approach of progressively learning the dynamic optimal return for each state-action pair using other variants of Deep Q-Networks including Double Q-Networks, Dueling Networks, etc.

\bibliography{references} 

\begin{thebibliography}{10}
\providecommand{\url}[1]{#1}
\csname url@samestyle\endcsname
\providecommand{\newblock}{\relax}
\providecommand{\bibinfo}[2]{#2}
\providecommand{\BIBentrySTDinterwordspacing}{\spaceskip=0pt\relax}
\providecommand{\BIBentryALTinterwordstretchfactor}{4}
\providecommand{\BIBentryALTinterwordspacing}{\spaceskip=\fontdimen2\font plus
\BIBentryALTinterwordstretchfactor\fontdimen3\font minus
  \fontdimen4\font\relax}
\providecommand{\BIBforeignlanguage}[2]{{%
\expandafter\ifx\csname l@#1\endcsname\relax
\typeout{** WARNING: IEEEtran.bst: No hyphenation pattern has been}%
\typeout{** loaded for the language `#1'. Using the pattern for}%
\typeout{** the default language instead.}%
\else
\language=\csname l@#1\endcsname
\fi
#2}}
\providecommand{\BIBdecl}{\relax}
\BIBdecl

\bibitem{wise2006uav}
R.~Wise and R.~Rysdyk, ``Uav coordination for autonomous target tracking,'' in
  \emph{AIAA Guidance, Navigation, and Control Conference and Exhibit}, 2006,
  p. 6453.

\bibitem{choi2014uav}
H.~Choi and Y.~Kim, ``Uav guidance using a monocular-vision sensor for aerial
  target tracking,'' \emph{Control Engineering Practice}, vol.~22, pp. 10--19,
  2014.

\bibitem{oh2013rendezvous}
H.~Oh, S.~Kim, H.-S. Shin, B.~A. White, A.~Tsourdos, and C.~A. Rabbath,
  ``Rendezvous and standoff target tracking guidance using differential
  geometry,'' \emph{Journal of Intelligent \& Robotic Systems}, vol.~69, no.
  1-4, pp. 389--405, 2013.

\bibitem{regina2011uav}
N.~Regina and M.~Zanzi, ``Uav guidance law for ground-based target trajectory
  tracking and loitering,'' in \emph{2011 Aerospace Conference}.\hskip 1em plus
  0.5em minus 0.4em\relax IEEE, 2011, pp. 1--9.

\bibitem{chen2009tracking}
H.~Chen, K.~Chang, and C.~S. Agate, ``Tracking with uav using
  tangent-plus-lyapunov vector field guidance,'' in \emph{2009 12th
  International Conference on Information Fusion}.\hskip 1em plus 0.5em minus
  0.4em\relax IEEE, 2009, pp. 363--372.

\bibitem{theodorakopoulos2008strategy}
P.~Theodorakopoulos and S.~Lacroix, ``A strategy for tracking a ground target
  with a uav,'' in \emph{2008 IEEE/RSJ International Conference on Intelligent
  Robots and Systems}.\hskip 1em plus 0.5em minus 0.4em\relax IEEE, 2008, pp.
  1254--1259.

\bibitem{pothen2017curvature}
A.~A. Pothen and A.~Ratnoo, ``Curvature-constrained lyapunov vector field for
  standoff target tracking,'' \emph{Journal of Guidance, Control, and
  Dynamics}, vol.~40, no.~10, pp. 2729--2736, 2017.

\bibitem{shaferman2008cooperative}
V.~Shaferman and T.~Shima, ``Cooperative uav tracking under urban occlusions
  and airspace limitations,'' in \emph{AIAA Guidance, Navigation and Control
  Conference and Exhibit}, 2008, p. 7136.

\bibitem{cook2013intelligent}
K.~Cook, E.~Bryan, H.~Yu, H.~Bai, K.~Seppi, and R.~Beard, ``Intelligent
  cooperative control for urban tracking with unmanned air vehicles,'' in
  \emph{2013 International Conference on Unmanned Aircraft Systems
  (ICUAS)}.\hskip 1em plus 0.5em minus 0.4em\relax IEEE, 2013, pp. 1--7.

\bibitem{yu2014cooperative}
H.~Yu, K.~Meier, M.~Argyle, and R.~W. Beard, ``Cooperative path planning for
  target tracking in urban environments using unmanned air and ground
  vehicles,'' \emph{IEEE/ASME Transactions on Mechatronics}, vol.~20, no.~2,
  pp. 541--552, 2014.

\bibitem{zhao2019detection}
X.~Zhao, F.~Pu, Z.~Wang, H.~Chen, and Z.~Xu, ``Detection, tracking, and
  geolocation of moving vehicle from uav using monocular camera,'' \emph{IEEE
  Access}, vol.~7, pp. 101\,160--101\,170, 2019.

\bibitem{watanabe2010optimal}
Y.~Watanabe and P.~Fabiani, ``Optimal guidance design for uav visual target
  tracking in an urban environment,'' \emph{IFAC Proceedings Volumes}, vol.~43,
  no.~15, pp. 69--74, 2010.

\bibitem{semsch2009autonomous}
E.~Semsch, M.~Jakob, D.~Pavlicek, and M.~Pechoucek, ``Autonomous uav
  surveillance in complex urban environments,'' in \emph{2009 IEEE/WIC/ACM
  International Joint Conference on Web Intelligence and Intelligent Agent
  Technology}, vol.~2.\hskip 1em plus 0.5em minus 0.4em\relax IEEE, 2009, pp.
  82--85.

\bibitem{ramirez2015urban}
J.-P. Ramirez-Paredes, E.~A. Doucette, J.~W. Curtis, and N.~R. Gans, ``Urban
  target search and tracking using a uav and unattended ground sensors,'' in
  \emph{2015 American Control Conference (ACC)}.\hskip 1em plus 0.5em minus
  0.4em\relax IEEE, 2015, pp. 2401--2407.

\bibitem{kim2010uav}
J.~Kim and J.~L. Crassidis, ``Uav path planning for maximum visibility of
  ground targets in an urban area,'' in \emph{2010 13th International
  Conference on Information Fusion}.\hskip 1em plus 0.5em minus 0.4em\relax
  IEEE, 2010, pp. 1--7.

\bibitem{wu2018path}
J.~Wu, H.~Wang, N.~Li, P.~Yao, Y.~Huang, and H.~Yang, ``Path planning for
  solar-powered uav in urban environment,'' \emph{Neurocomputing}, vol. 275,
  pp. 2055--2065, 2018.

\bibitem{theodorakopoulos2009uav}
P.~Theodorakopoulos and S.~Lacroix, ``Uav target tracking using an adversarial
  iterative prediction,'' in \emph{2009 IEEE International Conference on
  Robotics and Automation}.\hskip 1em plus 0.5em minus 0.4em\relax IEEE, 2009,
  pp. 2866--2871.

\bibitem{zhang2018coarse}
W.~Zhang, K.~Song, X.~Rong, and Y.~Li, ``Coarse-to-fine uav target tracking
  with deep reinforcement learning,'' \emph{IEEE Transactions on Automation
  Science and Engineering}, vol.~16, no.~4, pp. 1522--1530, 2018.

\bibitem{mueller2016benchmark}
M.~Mueller, N.~Smith, and B.~Ghanem, ``A benchmark and simulator for uav
  tracking,'' in \emph{European conference on computer vision}.\hskip 1em plus
  0.5em minus 0.4em\relax Springer, 2016, pp. 445--461.

\bibitem{Fujimoto2018AddressingFA}
S.~Fujimoto, H.~van Hoof, and D.~Meger, ``Addressing function approximation
  error in actor-critic methods,'' \emph{ArXiv}, vol. abs/1802.09477, 2018.

\bibitem{Hasselt2015DeepRL}
H.~van Hasselt, A.~Guez, and D.~Silver, ``Deep reinforcement learning with
  double q-learning,'' in \emph{AAAI}, 2015.

\bibitem{Mnih2013PlayingAW}
V.~Mnih, K.~Kavukcuoglu, D.~Silver, A.~Graves, I.~Antonoglou, D.~Wierstra, and
  M.~A. Riedmiller, ``Playing atari with deep reinforcement learning,''
  \emph{ArXiv}, vol. abs/1312.5602, 2013.

\bibitem{Wang2015DuelingNA}
Z.~Wang, T.~Schaul, M.~Hessel, H.~van Hasselt, M.~Lanctot, and N.~de~Freitas,
  ``Dueling network architectures for deep reinforcement learning,'' in
  \emph{ICML}, 2015.

\bibitem{Osband2016DeepEV}
I.~Osband, C.~Blundell, A.~Pritzel, and B.~V. Roy, ``Deep exploration via
  bootstrapped dqn,'' in \emph{NIPS}, 2016.

\bibitem{Hausknecht2015DeepRQ}
M.~J. Hausknecht and P.~Stone, ``Deep recurrent q-learning for partially
  observable mdps,'' in \emph{AAAI Fall Symposia}, 2015.

\end{thebibliography}
\bibliographystyle{IEEEtran}

\end{document}